\documentclass{article} % For LaTeX2e

\usepackage{hyperref}
\usepackage{url}

\usepackage{fp,xcolor,colortbl}
\usepackage{algorithm}
\usepackage{algorithmic}
\usepackage{amssymb}
\usepackage{booktabs}
\usepackage{lmodern}
\usepackage{mathtools}
\usepackage{nth}
\usepackage{graphicx}
\usepackage{subfigure}
\usepackage{iclr2016_conference,times}
\makeindex

\def\E{{\bf E}}

\def\0{{\bf 0}}
\def\1{{\bf 1}}

\newcommand{\he}[2][2]{%The first [2] means there are two arguments. The second
    \FPeval{\resa}{#2^#1}%
    \edef\processme{\noexpand\cellcolor[gray]{\resa}}%
    \processme
    #2%
}

\newcommand{\eqnref}[1]{\hyperref[formula:#1]{Eqn.~\ref*{formula:#1}}}

\def\index{\operatorname{index}}

\title{DoReFa-Net: Training Low Bitwidth Convolutional Neural Networks with Low
Bitwidth Gradients}

\author{Shuchang Zhou, Yuxin Wu, Zekun Ni, Xinyu Zhou, He Wen, Yuheng Zou \\
  Megvii Inc. \\
\texttt{\{zsc, wyx, nzk, zxy, wenhe, zouyuheng\}@megvii.com} \\
}

\begin{document}
\maketitle

\begin{abstract}
We propose DoReFa-Net, a method to train convolutional neural networks that have
low bitwidth weights and activations using low bitwidth parameter gradients. In
particular, during backward pass, parameter gradients are stochastically quantized to low bitwidth
numbers before being propagated to convolutional layers. As
convolutions during forward/backward passes can now operate on low bitwidth weights
and activations/gradients respectively, DoReFa-Net can use bit convolution
kernels to accelerate both training and inference. Moreover, as bit convolutions
can be efficiently implemented on CPU, FPGA, ASIC and GPU, DoReFa-Net opens the
way to accelerate training of low bitwidth neural network on these hardware.
Our experiments on SVHN and ImageNet datasets prove that DoReFa-Net can achieve comparable
prediction accuracy as 32-bit counterparts. For example, a DoReFa-Net derived
from AlexNet that has 1-bit weights, 2-bit activations, can be trained from
scratch using 6-bit gradients to get 46.1\% top-1 accuracy on ImageNet
validation set. The DoReFa-Net AlexNet model is released publicly.
\end{abstract}
% TODO
% We also release a bit convolution kernel written in ARM Neon assembly.
% The kernel is 10x faster than 32-bit counterparts, which may be used in
% scenarios like on-device machine learning.

\section{Introduction}
Recent progress in deep Convolutional Neural Networks (DCNN) has
considerably changed the landscape of computer vision \citep{krizhevsky2012imagenet}, speech
recognition \citep{hinton2012deep} and NLP \citep{bahdanau2014neural}.

However, a state-of-the-art DCNN usually has a lot of
parameters and high computational complexity, which both impedes its application in
embedded devices and slows down the iteration of its research and development.

For example, the training process of a DCNN may take up to weeks on a modern
multi-GPU server for large datasets like ImageNet \citep{deng2009imagenet}.
In light of this, substantial research efforts are invested in speeding
up DCNNs at both run-time and training-time, on both general-purpose
\citep{vanhoucke2011improving, gong2014compressing, han2015learning} and
specialized computer hardware \citep{farabet2011large, pham2012neuflow,
chen2014diannao, chen2014dadiannao}. Various approaches like quantization
\citep{wu2015quantized} and sparsification \citep{han2015deep} have also been
proposed.

Recent research efforts \citep{courbariaux2014training, kim2016bitwise,
rastegari2016xnor, merolla2016deep} have considerably reduced
both model size and computation complexity by using low bitwidth weights and low bitwidth activations. In particular, in BNN
\citep{courbariaux2016binarynet} and XNOR-Net \citep{rastegari2016xnor},
both weights and input activations of convolutional layers\footnote{Note
fully-connected layers are special cases of convolutional layers.} are binarized.
Hence during the forward pass the most computationally expensive convolutions can be
done by bitwise operation kernels, thanks to the following formula which
computes the dot product of two bit vectors $\mathbf{x}$ and $\mathbf{y}$ using
bitwise operations, where $bitcount$ counts the number of bits in a bit vector:
\begin{align}
\label{formula:bit-conv-kernel}
\mathbf{x} \cdot \mathbf{y} = 
\operatorname{bitcount}(\operatorname{and}(\mathbf{x}, \mathbf{y}))
\text{, } x_i, y_i \in \{0, 1\} \, \forall i \text{.}
\end{align}

\footnote{When $\mathbf{x}$ and $\mathbf{y}$ are vectors of $\{-1,
1\}$, \eqnref{bit-conv-kernel} has a variant that uses $xnor$
instead:
\begin{align}
\label{formula:bit-conv-kernel-xnor}
\mathbf{x} \cdot \mathbf{y} = 
N - 2\times\operatorname{bitcount}(\operatorname{xnor}(\mathbf{x}, \mathbf{y}))
\text{, } x_i, y_i \in \{-1, 1\} \, \forall i \text{.}
\end{align}
}

However, to the best of our knowledge, no previous work has succeeded in
quantizing gradients to numbers with bitwidth less than 8 during the backward
pass, while still achieving comparable prediction accuracy.
In some previous research \citep{gupta2015deep, courbariaux2014training}, convolutions involve at
least 10-bit numbers. In BNN and XNOR-Net, though weights are binarized,
gradients are in full precision, therefore the backward-pass
still requires convolution
between 1-bit numbers and 32-bit floating-points. The inability to exploit bit convolution during the backward pass means
that most training time of BNN and XNOR-Net will be spent in backward pass.

This paper makes the following contributions:
\begin{enumerate}
  \item We generalize the method of binarized neural networks to allow creating
  DoReFa-Net, a CNN that has arbitrary bitwidth in weights, activations, and
  gradients.
  As convolutions during forward/backward passes can then operate on low
  bit weights and activations/gradients respectively, DoReFa-Net can use bit convolution
kernels to accelerate both the forward pass and the backward pass of the
training process.
  \item As bit convolutions can be efficiently implemented on CPU, FPGA,
  ASIC and GPU, DoReFa-Net opens the way to accelerate low bitwidth neural
  network training on these hardware. In particular, with the power
  efficiency of FPGA and ASIC, we may considerably reduce energy consumption of
  low bitwidth neural network training.
  \item We explore the configuration space of bitwidth for
  weights, activations and gradients for DoReFa-Net. E.g., training a network
  using 1-bit weights, 1-bit activations and 2-bit gradients can lead to 93\%
   accuracy on SVHN dataset. In our experiments, gradients in general require
   larger bitwidth than activations, and activations in general require larger bitwidth
   than weights, to lessen the degradation of prediction accuracy compared to
   32-bit precision counterparts.
   We name our method ``DoReFa-Net'' to take note of these phenomena.
  \item We release in TensorFlow \citep{abaditensorflow} format a DoReFa-Net
\footnote{The model and supplement materials are available at
\url{https://github.com/ppwwyyxx/tensorpack/tree/master/examples/DoReFa-Net}}
derived from AlexNet \citep{krizhevsky2012imagenet} that gets 46.1\% in
single-crop top-1 accuracy on ILSVRC12 validation set. A reference implementation for training of a
DoReFa-net on SVHN dataset is also available.
\end{enumerate}

\section{DoReFa-Net}
In this section we detail our formulation of DoReFa-Net, a method to train
neural network that has low bitwidth weights, activations with
low bitwidth parameter gradients.
We note that while weights and activations can be deterministically quantized,
gradients need to be stochastically quantized.

We first outline how to exploit bit convolution kernel in DoReFa-Net and
then elaborate the method to quantize weights, activations and gradients to low
bitwidth numbers.

\subsection{Using Bit Convolution Kernels in Low Bitwidth Neural Network}
The 1-bit dot product kernel specified in \eqnref{bit-conv-kernel}
can also be used to compute dot product, and consequently convolution, for low bitwidth fixed-point integers.
Assume $\mathbf{x}$ is a sequence of $M$-bit fixed-point integers s.t.
$\mathbf{x} = \sum_{m=0}^{M-1} c_m(\mathbf{x}) 2^m$ and $\mathbf{y}$ is a
sequence of $K$-bit fixed-point integers s.t. $\mathbf{y} = \sum_{k=0}^{K-1}
c_k(\mathbf{y}) 2^k$ where $(c_m(\mathbf{x}))_{m=0}^{M-1}$ and
$(c_k(\mathbf{y}))_{k=0}^{K-1}$ are bit vectors, the dot product of $\mathbf{x}$
and $\mathbf{y}$ can be computed by bitwise operations as:
\begin{align}
\label{formula:bit-conv-kernel-integer}
\mathbf{x}\cdot \mathbf{y} = \sum_{m=0}^{M-1} \sum_{k=0}^{K-1} 2^{m+k} \,
\operatorname{bitcount}[\operatorname{and}(c_m(\mathbf{x}), c_k(\mathbf{y}))]
\text{, } \\ c_m(\mathbf{x})_i, c_k(\mathbf{y})_i \in \{0, 1\} \, \forall
i,\,m,\,k
\text{.}
\end{align}
% TODO c_m(x)_i is too weird

In the above equation, the computation complexity is
$O(MK)$, i.e., directly proportional to bitwidth of $\mathbf{x}$ and
$\mathbf{y}$.

% In the special case when $\mathbf{y}$ is composed of a bit string of $\{-1,
% 1\}$, we can exploit the following identity to exploit \eqnref{bit-conv-kernel}:
% \begin{align}
% \mathbf{x} \cdot \mathbf{y} = 2 \times (\mathbf{x} \cdot \frac{\mathbf{y} +
% 1}{2} - \mathbf{x})\text{, }x_i\in\{0, 1\}\text{, }y_i\in\{-1, 1\}\forall i.
% \end{align}

\subsection{Straight-Through Estimator}

The set of real numbers representable by a low bitwidth number $k$ only has a
small ordinality $2^k$.
However, mathematically any continuous function whose range is a small finite set would necessarily
always have zero gradient with respect to its input. We adopt the ``straight-through
estimator'' (STE) method \citep{hinton2012neural, bengio2013estimating} to
circumvent this problem. An STE can be thought of as an operator that
has arbitrary forward and backward operations.

A simple example is the STE defined for Bernoulli sampling with probability $p\in[0,1]$:
\begin{align*}
\textbf{Forward: }& q \sim Bernoulli(p) \\
\textbf{Backward: }& \frac{\partial{c}}{\partial{p}} = \frac{\partial{c}}{\partial{q}} \text{.}
\end{align*}

Here $c$ denotes the objective function. As sampling from a Bernoulli
distribution is not a differentiable function, ``$\frac{\partial{q}}{\partial{p}}$''
is not well defined, hence the backward pass cannot be directly constructed from the forward pass
using chain rule. Nevertheless, because
$q$ is on expectation equal to $p$, we may use the well-defined
gradient $\frac{\partial{c}}{\partial{q}}$ as an approximation for
$\frac{\partial{c}}{\partial{p}}$ and construct a STE as above. In other words,
STE construction gives a custom-defined ``$\frac{\partial{q}}{\partial{p}}$''.

An STE we will use extensively in this work is $\operatorname{\textbf{quantize}}_\mathbf{k}$ that quantizes a real number input $r_i\in[0, 1]$
to a $k$-bit number output $r_o \in  [0,1]$. This STE is defined as below:
\begin{align}
\label{formula:quantize-k}
\textbf{Forward: } &r_o = \frac{1}{2^k-1} \operatorname{round}((2^k-1) r_i)
\\
\textbf{Backward: }& \frac{\partial{c}}{\partial{r_i}} = \frac{\partial{c}}{\partial{r_o}} \text{.}
\end{align}

It is obvious by construction that the output $q$ of $\operatorname{quantize}_k$
STE is a real number representable by $k$ bits.
Also, since $r_o$ is a $k$-bit fixed-point integer,
the dot product of two sequences of such $k$-bit real numbers can be efficiently
calculated, by using fixed-point integer dot product in
\eqnref{bit-conv-kernel-integer} followed by proper scaling.

\subsection{Low Bitwidth Quantization of Weights}
\label{subsec:bin-weight}

In this section we detail our approach to getting low bitwidth weights.

In previous works, STE has been used to binarize the weights. For example
in BNN, weights are binarized by the following STE:
\begin{align*}
\textbf{Forward: }& r_o = \operatorname{sign}(r_i) \\
\textbf{Backward: }& \frac{\partial{c}}{\partial{r_i}} =
\frac{\partial{c}}{\partial{r_o}} \mathbb{I}_{|r_i|\le 1}
\text{.}
\end{align*}

Here $\operatorname{sign}(r_i) = 2 \mathbb{I}_{r_i \ge 0} - 1$ returns one of
two possible values: $\{-1,\,1\}$.

In XNOR-Net, weights are binarized by the following STE, with the difference being that
weights are scaled after binarized:
\begin{align*}
\textbf{Forward: } &r_o = \operatorname{sign}(r_i) \times \operatorname{\E}_{F}(|r_i|)
\\
\textbf{Backward: }& \frac{\partial{c}}{\partial{r_i}} =
\frac{\partial{c}}{\partial{r_o}} \text{.}
\end{align*}

In XNOR-Net, the scaling factor $\operatorname{\E}_{F}(|r_i|)$ is the mean of
absolute value of each output channel of weights.
The rationale is that introducing this scaling factor will increase the value
range of weights, while still being able to exploit bit convolution kernels.
However, the channel-wise scaling factors will make it impossible to exploit
bit convolution kernels when computing the convolution between gradients
and the weights during back propagation. Hence, in our experiments, we
use a constant scalar to scale all filters instead of doing channel-wise
scaling.
We use the following STE for all neural networks that have binary weights
in this paper:

\begin{align}
\label{formula:global-scale-weight}
\textbf{Forward: } &r_o = \operatorname{sign}(r_i) \times \operatorname{\E}(|r_i|)
\\
\textbf{Backward: } &\frac{\partial{c}}{\partial{r_i}} =
\frac{\partial{c}}{\partial{r_o}}
\text{.}
\end{align}

In case we use $k$-bit representation of the weights with $k>1$, we apply the
STE $f_{\omega}^k $ to weights as follows:
\begin{align}
\label{formula:low-bit-weight}
\textbf{Forward: } &r_o = f_{\omega}^k(r_i) =
2 \operatorname{quantize_k}(\frac{\tanh(r_i)}{2 \max(|\tanh(r_i)|)} + \frac12) - 1 \text{.}
\\
\textbf{Backward: } &\frac{\partial{c}}{\partial{r_i}} =
\frac{\partial{r_o}}{\partial{r_i}}\frac{\partial{c}}{\partial{r_o}} \, \,
\footnotemark
\end{align}
\footnotetext{Here $ \frac{\partial{r_o}}{\partial{r_i}}$ is well-defined because
  we already defined $ \operatorname{quantize}_k$ as an STE}

Note here we use $tanh$ to limit the value range of weights to $[-1, 1]$ before
quantizing to $k$-bit. By construction, $\frac{\tanh(r_i)}{2 \max(|\tanh(r_i)|)} + \frac12$ is
a number in $[0, 1]$, where the maximum is taken over all weights in that layer. $quantize_k$ will then quantize this number to $k$-bit
fixed-point ranging in $[0, 1]$. Finally an affine transform will bring the range of
$f_\omega^k(r_i)$ to $[-1, 1]$.

Note that when $k=1$, \eqnref{low-bit-weight} is different from
\eqnref{global-scale-weight}, providing
a different way of binarizing weights.
Nevertheless, we find this difference insignificant in experiments.

\subsection{Low Bitwidth Quantization of Activations}
Next we detail our approach to getting low bitwidth activations that are input
to convolutions, which is of critical importance in replacing floating-point
convolutions by less computation-intensive bit convolutions.

In BNN and XNOR-Net, activations are binarized in the same
way as weights. However, we fail to reproduce the results of XNOR-Net
if we follow their methods of binarizing activations, and the binarizing
approach in BNN is claimed by \citep{rastegari2016xnor} to cause severe
prediction accuracy degradation when applied on ImageNet models like AlexNet.
Hence instead, we apply an STE on input activations $r$ of each weight layer.
Here we assume the output of the previous layer has
passed through a bounded activation function $h$, which ensures $r \in [0, 1]$.
In DoReFa-Net, quantization of activations $r$ to $k$-bit is simply:
\begin{align}
\label{formula:low-bit-activation}
f_\alpha^k(r) = \operatorname{quantize_k}(r)\text{.}
\end{align}

% We observed that the choice of activation function $h$ has a profound impact on
% prediction accuracy. In our experiments we find following
% activations functions to be useful for some models:
% \begin{enumerate}
%   \item $h(x) = \frac{\tanh(x) + 1}{2}$
%   \item $h(x) = \operatorname{clip}(x, 0, 1)$
%   \item $h(x) = \min(1, |x|)$
% \end{enumerate}

\subsection{Low Bitwidth Quantization of Gradients}
We have demonstrated deterministic quantization to produce low bitwidth
weights and activations. However, we find stochastic quantization is necessary for
low bitwidth gradients to be effective. This is in
agreement with experiments of \citep{gupta2015deep} on 16-bit weights and
16-bit gradients.

To quantize gradients to low bitwidth, it is important to note that gradients
are unbounded and may have significantly larger value range than activations.
Recall in
\eqnref{low-bit-activation}, we can map the range of activations
to $[0, 1]$ by passing values through differentiable nonlinear functions.
However, this kind of construction does not exist for gradients.
Therefore we designed the following function for $k$-bit quantization of
gradients:
\begin{align*}
\tilde{f}_\gamma^k(\mathrm{d}r) = 2\operatorname{max}_0(|\mathrm{d}r|) \left
[\operatorname{quantize}_k(\frac{\mathrm{d}r} {2\operatorname{max}_0(|\mathrm{d}r|)} + \frac12) - \frac12\right ]
\text{.}
\end{align*}

Here $\mathrm{d}r=\frac{\partial c}{\partial r}$ is the back-propagated gradient of the output $r$ of some layer, and the maximum is taken over
all axis of the gradient tensor $\mathrm{d}r$ except for the mini-batch axis (therefore each instance in a mini-batch will have its own
scaling factor).
The above function first applies an affine transform on the gradient, to map it into $[0,1]$,
and then inverts the transform after quantization.

To further compensate the potential bias introduced by gradient quantization, we introduce
an extra noise function $N(k) = \frac{\sigma}{2^k-1}$ where $\sigma \sim
Uniform(-0.5, 0.5)$. \footnote{Note here we do not
need clip value of $N(k)$ as the two end points of a uniform
distribution are almost surely never attained.} The noise therefore has the same
magnitude as the possible quantization error.
We find that the artificial noise to be critical for achieving good
performance.
Finally, the expression we'll use to quantize gradients to $k$-bit numbers is as follows:
\begin{align}
\label{formula:low-bit-gradient}
f_\gamma^k(\mathrm{d}r) = 2\operatorname{max}_0(|\mathrm{d}r|) \left[
\operatorname{quantize}_k[\frac{\mathrm{d}r} {2\operatorname{max}_0(|\mathrm{d}r|)} + \frac12 + N(k)] - \frac12)\right]
\text{.}
\end{align}

The quantization of gradient is done on the backward pass only. Hence
we apply the following STE on the output of each convolution layer:
\begin{align}
\textbf{Forward: }& r_o = r_i
\\
\textbf{Backward: }& \frac{\partial{c}}{\partial{r_i}} =
f_\gamma^k(\frac{\partial{c}}{\partial{r_o}})
\text{.}
\end{align}

\begin{algorithm}
\caption{Training a $L$-layer DoReFa-Net with $W$-bit weights and $A$-bit activations
using $G$-bit gradients.
Weights, activations and gradients are quantized according to
\eqnref{low-bit-weight}, \eqnref{low-bit-activation},
\eqnref{low-bit-gradient}, respectively.
}
\label{alg:train-dorefa}
\begin{algorithmic}[1]
\REQUIRE{a minibatch of inputs and targets $(a_0, a^*)$, previous weights $W$, learning rate $\eta$}
\ENSURE{updated weights $W^{t+1}$}

\{1. Computing the parameter gradients:\}

\{1.1 Forward propagation:\}
\FOR{$k = 1\, \TO\, L $}
\STATE{$W_k^b \gets f_\omega^W(W_k)$}
\STATE{$\tilde{a}_k \gets \texttt{forward}(a_{k-1}^b,W_k^b)$}
\STATE{$a_k \gets h(\tilde{a}_k)$}
\IF{$k < L$}
\STATE{$a_k^b \gets f_\alpha^A(a_k)$}
\ENDIF
\STATE{Optionally apply pooling}
\ENDFOR

\{1.2 Backward propagation:\}

%\{Note that the gradients $g^b_{a_k}$ are of $G$-bit.\}

Compute $g_{a_L} = \frac{\partial C}{\partial a_L}$
knowing $a_L$ and $a^*$.

\FOR{$k = L\, \TO\, 1 $}
\STATE{Back-propagate $ g_{a_k}$ through activation function $h$}
\STATE{$g^b_{a_k}\gets f_\gamma^G(g_{a_k})$}
\STATE{$g_{a_{k-1}} \gets \texttt{backward\_input}(g^b_{a_k}, W_k^b)$}
\STATE{$g_{W_{k}^b} \gets \texttt{backward\_weight}(g^b_{a_k}, a_{k-1}^b)$}
\STATE{Back-propagate gradients through pooling layer if there is one}
\ENDFOR

\{2. Accumulating the parameters gradients:\}
\FOR{$k = 1\, \TO\, L $}
\STATE{$ g_{W_k} = g_{W_k^b} \frac{\partial W_k^b}{\partial W_k}$}
\STATE{$W_k^{t+1} \gets Update(W_k, g_{W_k}, \eta)$}
\ENDFOR

\end{algorithmic}
\end{algorithm}

\subsection{The Algorithm for DoReFa-Net}
We give a sample training algorithm of DoReFa-Net as
Algorithm~\ref{alg:train-dorefa}. W.l.o.g., the network is assumed to have a
feed-forward linear topology, and details like batch normalization and pooling layers are
omitted. Note that all the expensive operations \texttt{forward, backward\_input, backward\_weight},
in convolutional as well as fully-connected layers,
are now operating on low bitwidth numbers.
By construction, there is always an affine mapping between these low bitwidth numbers and fixed-point integers.
As a result, all the expensive operations can be accelerated significantly by the
fixed-point integer dot product kernel
(\eqnref{bit-conv-kernel-integer}).

\subsection{First and the last layer}
Among all layers in a DCNN, the first and the last layers appear to be different
from the rest, as they are interfacing the input and output of the network. For
the first layer, the input is often an image, which may contain 8-bit features.
On the other hand, the output layer typically produce approximately one-hot
vectors, which are close to bit vectors by definition.
It is an interesting question whether these differences would cause the first and last layer to exhibit different behavior
when converted to low bitwidth counterparts.

In the related work of \citep{han2015learning} which converts network weights to
sparse tensors, introducing the same ratio of zeros in the first
convolutional layer is found to cause more prediction accuracy degradation than
in the other convolutional layers. Based on this intuition as well as the observation that
the inputs to the first layer often contain only a few channels
and constitutes a small proportion of total computation complexity, we perform
most of our experiments by not quantizing the
weights of the first convolutional layer, unless noted
otherwise. Nevertheless, the outputs of the first convolutional layer are
quantized to low bitwidth as they would be used by the consequent convolutional layer.

Similarly, when the output number of class is small, to stay away from
potential degradation of prediction accuracy, we leave the last fully-connected layer
intact unless noted otherwise.
Nevertheless, the gradients back-propagated from the final FC layer are properly
quantized.

We will give the empirical evidence in
Section~\ref{subsec:first-and-last-layer}.

% \subsection{Periodical Synchronization of STE}
% One of the major motivation of STE is to have different values for forward
% propagation and accumulation of gradients.
% For example, when weights are converted to low bitwidth numbers by
% \ref{func:low-bit-weight}, the value used to compute activations during forward
% pass are of low bitwidth, but the weights accumulating gradients are usually of
% 32-bit precision. Hence there will be a developing discrepancy between the
% weights and their low bitwidth form, potentially leading to less accurate
% activations.
%
% We introduce a periodical synchronization operation for STE. I.e., every a few
% epochs, we will set the weights to be the same as its low bitwidth form. Experiments
% find that this can improve the final accuracy by a few percentage.
%
% XXX

\subsection{Reducing Run-time Memory Footprint by Fusing Nonlinear Function and
Rounding} One of the motivations for creating low bitwidth neural network is to save
run-time memory footprint in inference.
A naive implementation of Algorithm~\ref{alg:train-dorefa} would store
activations $h(a_k)$ in full-precision numbers,
consuming much memory during run-time.
In particular, if $h$ involves floating-point arithmetics,
there will be non-negligible amount of non-bitwise operations related to
computations of $h(a_k)$.

There are simple solutions to this problem.
Notice that it is possible to fuse Step~3, Step~4, Step~6 to avoid
storing intermediate results in full-precision.
Apart from this, when $h$ is monotonic, $f_\alpha \cdot h$ is also monotonic, the few
possible values of $a_k^b$ corresponds to several non-overlapping value ranges
of $a_k$, hence we can implement computation of $a_k^b = f_\alpha(h(a_k))$ by several
comparisons between fixed point numbers and avoid generating intermediate
results.

Similarly, it would also be desirable to fuse Step~11 $ \sim$ Step~12, and Step~13 of previous
iteration to avoid generation and storing of $g_{a_k}$. The situation would be
more complex when there are intermediate pooling layers. Nevertheless, if the pooling layer is
max-pooling, we can do the
fusion as $\operatorname{quantize}_k$ function commutes with $\max$ function:
\begin{align}
\operatorname{quantize}_k(\max(a, b)) = \max(\operatorname{quantize}_k(a),
\operatorname{quantize}_k(b))) \text{,}
\end{align}

hence again $g_{a_k}^b$ can be generated from $g_{a_k}$ by comparisons between
fixed-point numbers.

\section{Experiment Results}
\subsection{Configuration Space Exploration}
\begin{table}[!ht] \centering \small
\caption{Comparison of prediction accuracy for SVHN with different choices of
Bit-width in a DoReFa-Net. $W$, $A$, $G$ are bitwidths of weights, activations
and gradients respectively. When bitwidth is 32, we simply remove the
quantization functions.}
\begin{center}
\begin{tabular}{p{0.02\linewidth} p{0.02\linewidth} p{0.02\linewidth}
p{0.12\linewidth} p{0.12\linewidth} p{0.08\linewidth} p{0.09\linewidth}
p{0.09\linewidth} p{0.09\linewidth}  p{0.09\linewidth}}
\toprule  W & A & G & Training Complexity &
Inference Complexity & Storage Relative Size & Model A Accuracy & Model B
Accuracy & Model C Accuracy  & Model D Accuracy\\
    \midrule 1 & 1 & 2 & 3 & 1 & 1 & \he{0.934} & \he{0.924} & \he{0.910} &
    \he{0.803} \\
    \midrule 1 & 1 & 4 & 5 & 1 & 1 & \he{0.968} & \he{0.961} & \he{0.916} &
    \he{0.846} \\
    \midrule 1 & 1 & 8 & 9 & 1 & 1 & \he{0.970} & \he{0.962} & \he{0.902} &
    \he{0.828} \\
    \midrule 1 & 1 & 32 & - & - & 1 & \he{0.971} & \he{0.963} & \he{0.921} &
    \he{0.841} \\
    \hline
    \midrule 1 & 2 & 2 & 4 & 2 & 1 & \he{0.909} & \he{0.930} & \he{0.900} &
    \he{0.808} \\
    \midrule 1 & 2 & 3 & 5 & 2 & 1 & \he{0.968} & \he{0.964} & \he{0.934} &
    \he{0.878} \\
    \midrule 1 & 2 & 4 & 6 & 2 & 1 & \he{0.975} & \he{0.969} & \he{0.939} &
    \he{0.878}
    \\
    \midrule 2 & 1 & 2 & 6 & 2 & 2 & \he{0.927} & \he{0.928} & \he{0.909} &
    \he{0.846}\\
    \midrule 2 & 1 & 4 & 10 & 2 & 2 & \he{0.969} & \he{0.957} & \he{0.904} &
    \he{0.827} \\
    \midrule 1 & 2 & 8 & 10 & 2 & 1 & \he{0.975} & \he{0.971} & \he{0.946} &
    \he{0.866}
    \\
    \midrule 1 & 2 & 32 & - & - & 1 & \he{0.976} & \he{0.970} & \he{0.950} &
    \he{0.865} \\
        \hline
    \midrule 1 & 3 & 3 & 6 & 3 & 1 & \he{0.968} & \he{0.964} & \he{0.946}  &
    \he{0.887} \\
    \midrule 1 & 3 & 4 & 7 & 3 & 1 & \he{0.974} & \he{0.974} & \he{0.959}  &
    \he{0.897}
    \\
    \midrule 1 & 3 & 6 & 9 & 3 & 1 & \he{0.977} & \he{0.974} & \he{0.949}  &
    \he{0.916} \\
        \hline
    \midrule 1 & 4 & 2 & 6 & 4 & 1 & \he{0.815} & \he{0.898} & \he{0.911} &
    \he{0.868} \\
    \midrule 1 & 4 & 4 & 8 & 4 & 1 & \he{0.975} & \he{0.974} & \he{0.962} &
    \he{0.915} \\
    \midrule 1 & 4 & 8 & 12 & 4 & 1 & \he{0.977} & \he{0.975} & \he{0.955} &
    \he{0.895} \\
    \midrule 2 & 2 & 2 & 8 & 4 & 1 & \he{0.900} & \he{0.919}  & \he{0.856} &
    \he{0.842} \\
    \hline
    \midrule 8 & 8 & 8 & - & - & 8 & &  & \he{0.970} & \he{0.955} \\
    \hline    
    \midrule 32 & 32 & 32 & - & - & 32 & \he{0.975} & \he{0.975} & \he{0.972} &
    \he{0.950}
    \\
\hline
\end{tabular}
\end{center} \label{tab:svhn_wag}
\end{table}

We explore the configuration space of combinations of bitwidth of weights,
activations and gradients by experiments on the SVHN dataset.

The SVHN dataset \citep{netzer2011reading}
is a real-world digit recognition dataset consisting of photos of house numbers in Google Street View images.
We consider the ``cropped'' format of the dataset: 32-by-32 colored images centered
around a single character. There are 73257 digits for training, 26032 digits
for testing, and 531131 less difficult samples which can be used as extra training
data. The images are resized to 40x40 before fed into network.

For convolutions in a DoReFa-Net, if we have $W$-bit weights, $A$-bit activations
and $G$-bit gradients, the relative forward and backward computation complexity,
storage relative size, can be computed from \eqnref{bit-conv-kernel-integer} and we list them in
Table~\ref{tab:svhn_wag}. As it would not be computationally efficient to use
bit convolution kernels for convolutions between 32-bit numbers, and noting that
previous works like BNN and XNOR-net have already compared bit convolution
kernels with 32-bit convolution kernels, we will omit the complexity comparison of computation complexity for
the 32-bit control experiments.

We use the prediction accuracy of several CNN models on SVHN dataset to evaluate
the efficacy of configurations. Model A is a CNN that costs about 80 FLOPs for one
40x40 image, and it consists of seven convolutional layers and one fully-connected
layer.

Model B, C, D is derived from Model A by reducing the number of channels for all
seven convolutional layers by 50\%, 75\%, 87.5\%, respectively.
The listed prediction accuracy is the maximum accuracy on test set over 200
epochs.
We use ADAM \citep{kingma2014adam} learning rule with 0.001 as learning rate.

In general, having low bitwidth weights, activations and gradients will cause
degradation in prediction accuracy. But it should be noted that low bitwidth networks
will have much reduced resource requirement.

As balancing between multiple factors like training time, inference time, model
size and accuracy is more a problem of practical trade-off, there will be no
definite conclusion as which combination of $(W,\,A,\,G)$ one should choose.
Nevertheless, we find in these experiments that weights, activations and
gradients are progressively more sensitive to bitwidth, and using gradients
with $ G\le 4$ would significantly degrade prediction accuracy. Based on these
observations, we take $(W,\,A)$ = $(1,\,2)$ and $G\ge 4$ as rational
combinations and use them for most of our experiments on ImageNet dataset.

% Interestingly, By comparing row $(W,\,A,\,G)$ = $(1,\,2,\,32)$ with row $(W,\,A,\,G)$ =
% $(1,\,2,\,4)$, it can also be seen that for low bitwidth networks, training with low
% bit gradients may boost prediction accuracy.
% The credit may belong to the regularization effect introduced by the noise in gradients, similar to
% previous works in adding noise to gradients \citep{neelakantan2015adding}.

Table~\ref{tab:svhn_wag} also shows that the relative number of channels
significantly affect the prediction quality degradation resulting from bitwidth
reduction. For example, there is
no significant loss of prediction accuracy when going from 32-bit model to
DoReFa-Net for Model A, which is not the case for Model C. We conjecture that
``more capable'' models like those with more channels will be less
sensitive to bitwidth differences. On the other hand, Table~\ref{tab:svhn_wag}
also suggests a method to compensate for the prediction quality degradation, by
increasing bitwidth of activations for models with less channels, at the cost
of increasing computation complexity for inference and training.
However, optimal bitwidth of gradient seems less related to model channel
numbers and prediction quality saturates with 8-bit gradients most of the time.

\subsection{ImageNet}
We further evaluates DoReFa-Net on ILSVRC12 \citep{deng2009imagenet} image
classification dataset, which contains about 1.2 million high-resolution natural images for training that spans
1000 categories of objects.
The validation set contains 50k images.
We report our single-crop evaluation result
using top-1 accuracy. The images are resized to 224x224
before fed into the network.

The results are listed in Table~\ref{tab:imagenet_wag}. The baseline AlexNet
model that scores 55.9\% single-crop top-1 accuracy is a best-effort replication of the model in \citep{krizhevsky2012imagenet},
with the second, fourth and fifth convolutions split into two parallel blocks.
We replace the Local Contrast Renormalization layer with Batch Normalization layer
\citep{ioffe2015batch}. We use ADAM learning rule with
learning rate $10^{-4}$ at the start, and later decrease learning rate to
$10^{-5}$ and consequently $10^{-6}$ when accuracy curves become flat.

From the table, it can be seen that increasing bitwidth of activation from 1-bit
to 2-bit and even to 4-bit, while still keep 1-bit weights, leads to significant
accuracy increase, approaching the accuracy of model where both weights
and activations are 32-bit. Rounding gradients to 6-bit produces similar
accuracies as 32-bit gradients, in experiments of ``1-1-6'' v.s.\ ``1-1-32'',
``1-2-6'' v.s.\ ``1-2-32'', and ``1-3-6'' v.s.\ ``1-3-32''.

The rows with ``initialized'' means the model training has been initialized with
a 32-bit model. It can be seen that there is a considerable gap between
the best accuracy of a trained-from-scratch-model and an initialized model.
Closing this gap is left to future work. Nevertheless, it show the potential in
improving accuracy of DoReFa-Net.

\begin{table}[!ht] \centering \small
\caption{Comparison of prediction accuracy for ImageNet with different choices
of bitwidth in a DoReFa-Net. $W$, $A$, $G$ are bitwidths of weights, activations
and gradients respectively. Single-crop top-1 accuracy is given.
Note the BNN result is
    reported by \citep{rastegari2016xnor}, not by original authors.
We do not quantize the first and last layers of AlexNet to low bitwidth, as BNN and
XNOR-Net do.}
\begin{center}
\begin{tabular}{p{0.02\linewidth} p{0.02\linewidth} p{0.02\linewidth}
p{0.13\linewidth} p{0.13\linewidth} p{0.1\linewidth} p{0.2\linewidth}
}
\toprule  W & A & G & Training Complexity &
Inference Complexity & Storage Relative Size & AlexNet Accuracy\\
%    \midrule 1 & 1 & 4 & 5 & 1 & 1 &  \\
    \midrule 1 & 1 & 6 & 7 & 1 & 1 & 0.395\\
    \midrule 1 & 1 & 8 & 9 & 1 & 1 & 0.395 \\
    \midrule 1 & 1 & 32 & - & 1 & 1 & 0.279 (BNN) \\
    \midrule 1 & 1 & 32 & - & 1 & 1 & 0.442 (XNOR-Net) \\
    \midrule 1 & 1 & 32 & - & 1 & 1 & 0.401 \\    
    \midrule 1 & 1 & 32 & - & 1 & 1 & 0.436 (initialized) \\
    \hline
    \midrule 1 & 2 & 6 & 8 & 2 & 1 & 0.461 \\
    \midrule 1 & 2 & 8 & 10 & 2 & 1 & 0.463 \\
    \midrule 1 & 2 & 32 & - & 2 & 1 & 0.477 \\
    \midrule 1 & 2 & 32 & - & 2 & 1 & 0.498 (initialized) \\
    \hline
    \midrule 1 & 3 & 6 & 9 & 3 & 1 & 0.471 \\
    \midrule 1 & 3 & 32 & - & 3 & 1 & 0.484 \\
    \hline
    \midrule 1 & 4 & 6 & - & 4 & 1 & 0.482 \\        
    \midrule 1 & 4 & 32 & - & 4 & 1 & 0.503 \\    
    \midrule 1 & 4 & 32 & - & 4 & 1 & 0.530 (initialized) \\
    \hline
    \midrule 8 & 8 & 8 & - & - & 8 & 0.530 \\        
    \hline
    \midrule 32 & 32 & 32 & - & - & 32 & 0.559 \\
\hline
\end{tabular}
\end{center}
\label{tab:imagenet_wag}
\end{table}

\subsubsection{Training curves}
%---------------------------------Figure---------------------------------%
\begin{figure}
\begin{center}
\includegraphics[height=0.3\textheight, width=0.7\textwidth]{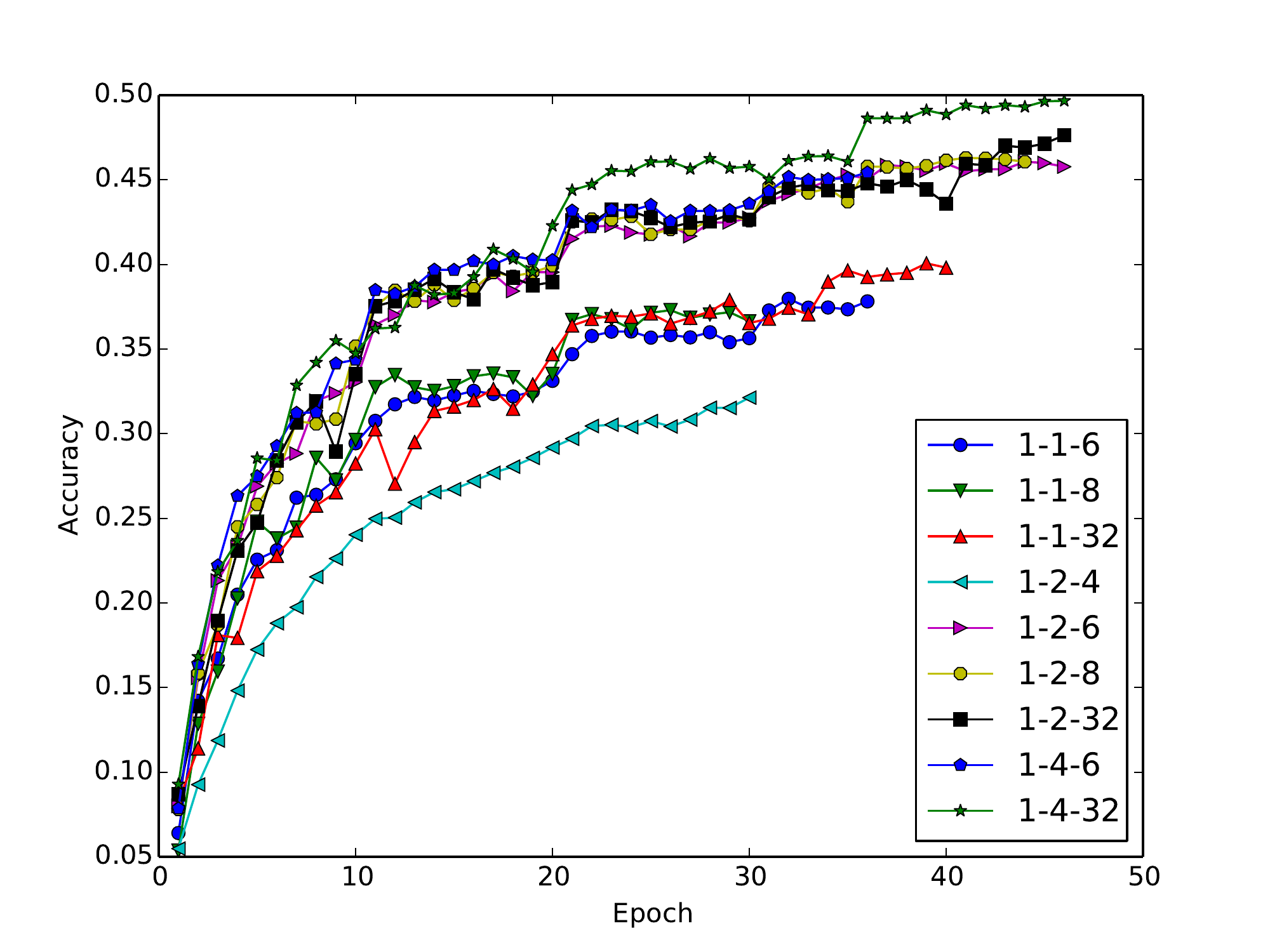}
\end{center}
   \caption{Prediction accuracy of AlexNet variants on Validation Set of
   ImageNet indexed by epoch number. ``W-A-G'' gives the specification of
   bitwidths of weights, activations and gradients.
   E.g., ``1-2-4'' stands for the case when weights are 1-bit, activations are
   2-bit and gradients are 4-bit. The figure is best viewed in color.}
\label{fig:dorefa-vs-32}
\end{figure}

Figure~\ref{fig:dorefa-vs-32} shows the evolution of accuracy v.s.\ epoch curves
of DoReFa-Net. It can be seen that quantizing gradients to be 6-bit does not
cause the training curve to be significantly different from not quantizing
gradients. However, using 4-bit gradients as in ``1-2-4'' leads to significant
accuracy degradation.

\subsubsection{Histogram of Weights, Activations and Gradients}
\begin{figure}[t!]
    \centering
    \begin{subfigure}[]
        \centering
        \includegraphics[height=0.3\textheight]{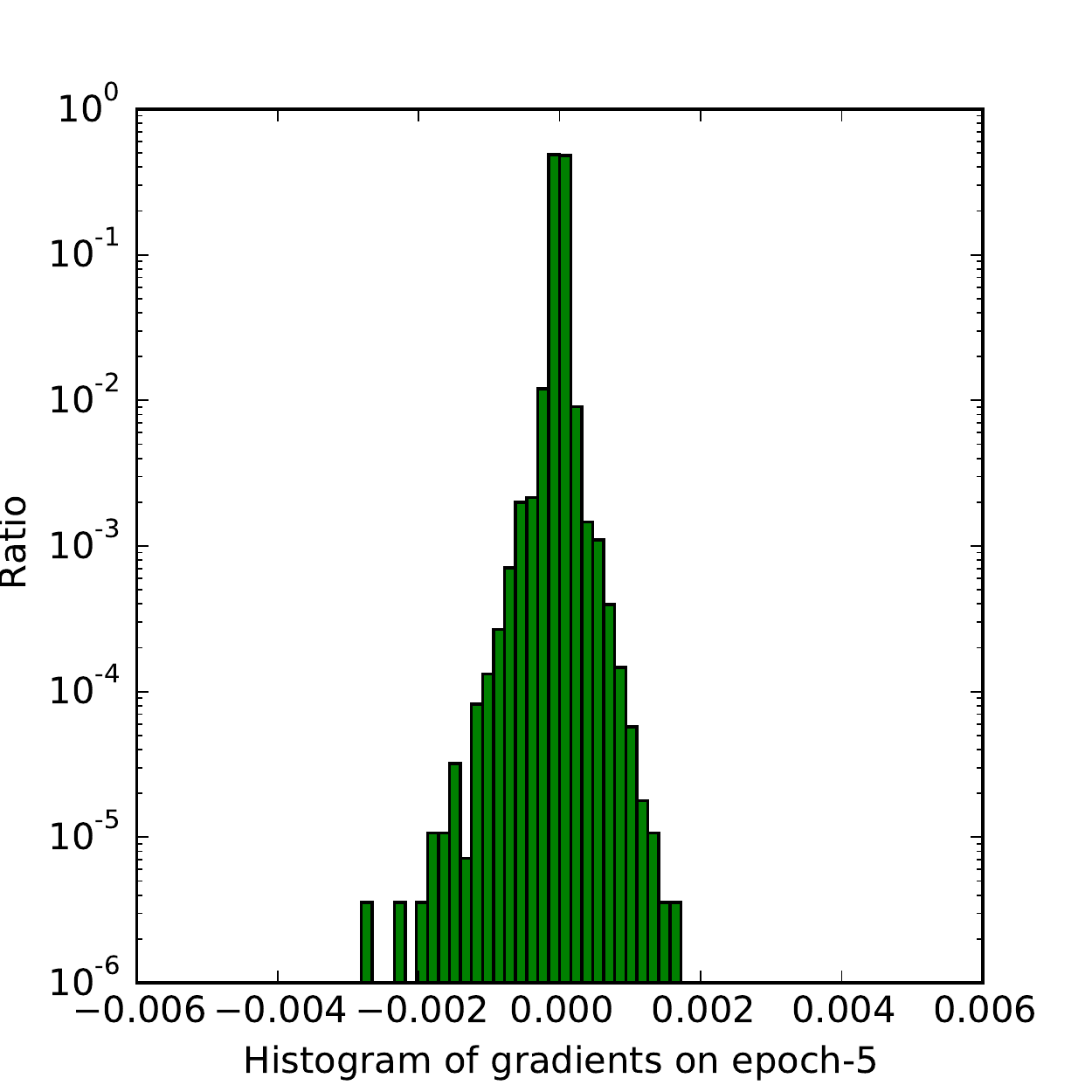}
        %\subcaption{Histogram of weights}
    \end{subfigure}%
    ~ 
    \begin{subfigure}[]
        \centering
        \includegraphics[height=0.3\textheight]{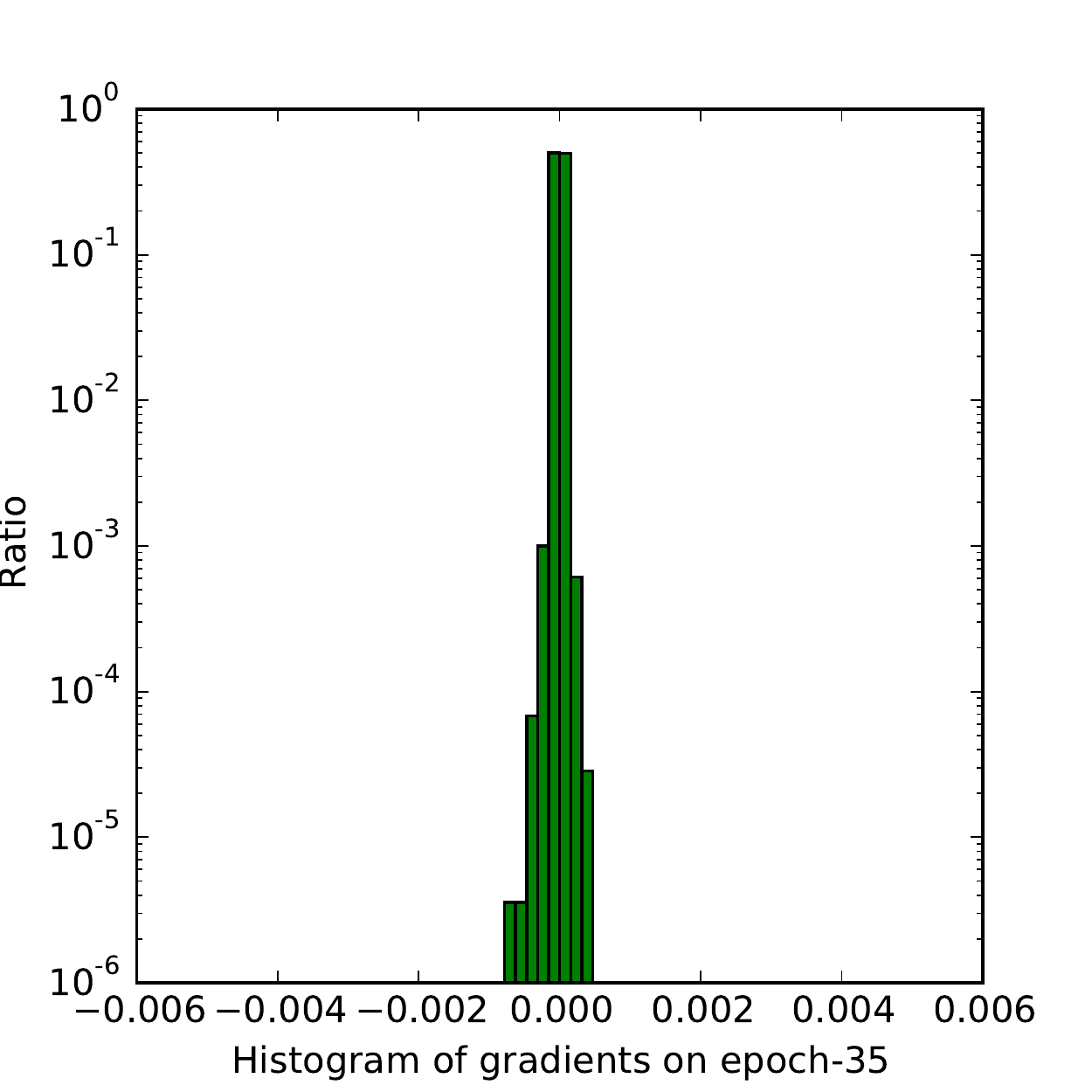}
        %\subcaption{Histogram of activations}
    \end{subfigure}
\caption{Histogram of gradients of layer ``conv3'' of ``1-2-6''
   AlexNet model at epoch 5 and 35.
   The y-axis is in logarithmic scale.}
\label{fig:gradient_distrib}
\end{figure}

Figure~\ref{fig:gradient_distrib} shows the histogram of gradients of layer
``conv3'' of ``1-2-6'' AlexNet model at epoch 5 and 35. As the histogram remains
mostly unchanged with epoch number, we omit the histograms of the other epochs
for clarity.

\begin{figure}[t!]
    \centering
    \begin{subfigure}[]
        \centering
        \includegraphics[height=0.3\textheight]{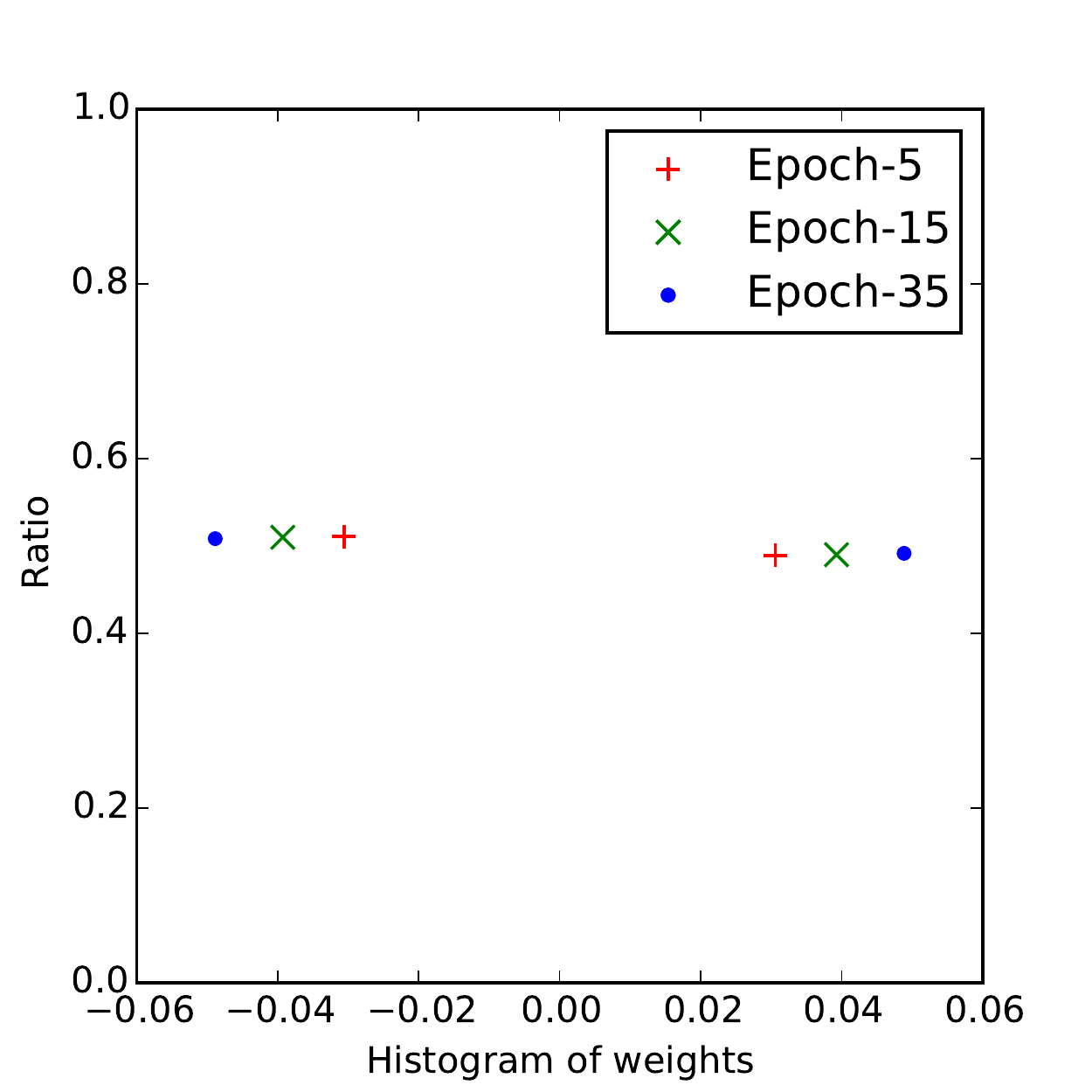}
        %\subcaption{Histogram of weights}
    \end{subfigure}%
    ~ 
    \begin{subfigure}[]
        \centering
        \includegraphics[height=0.3\textheight]{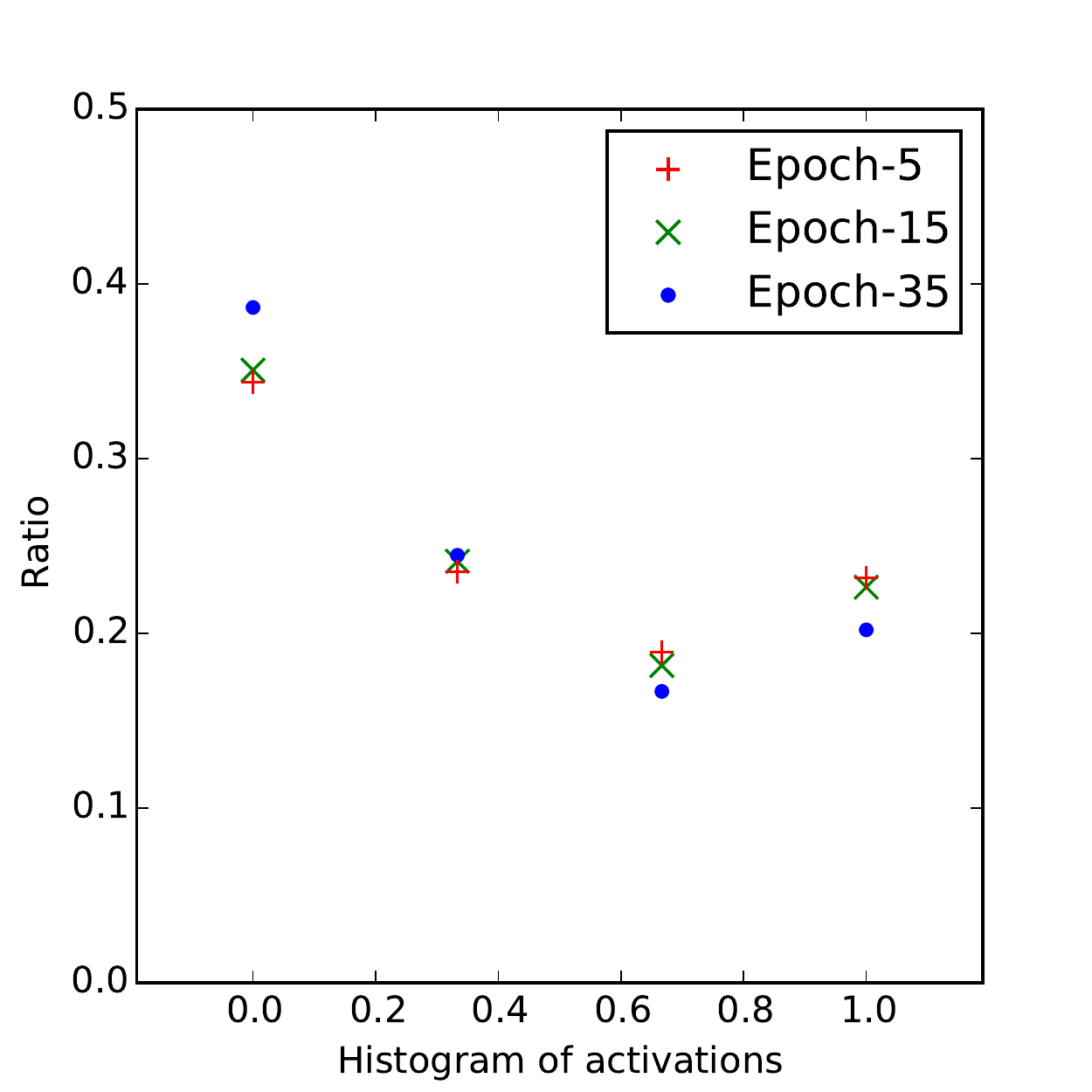}
        %\subcaption{Histogram of activations}
    \end{subfigure}
\caption{(a) is histogram of weights of layer ``conv3'' of ``1-2-6''
AlexNet model at epoch 5, 15 and 35. There are two possible values at a specific
   epoch since the weights are scaled 1-bit. (b) is histogram of activation
   of layer ``conv3'' of ``1-2-6'' AlexNet model at epoch 5, 15 and 35. There are four possible values at a specific
   epoch since the activations are 2-bit.}
\label{fig:weight_distrib}   
\end{figure}

Figure~\ref{fig:weight_distrib}(a) shows the histogram of weights of layer
``conv3'' of ``1-2-6'' AlexNet model at epoch 5, 15 and 35. Though the scale of
the weights changes with epoch number, the distribution of weights are
approximately symmetric.

Figure~\ref{fig:weight_distrib}(b) shows the histogram of activations of layer
``conv3'' of ``1-2-6'' AlexNet model at epoch 5, 15 and 35. The distributions of
activations are stable throughout the training process.

%---------------------------------Figure---------------------------------%
% \begin{figure}
% \begin{center}
% \includegraphics[height=0.3\textheight,
% width=0.6\textwidth]{gradients}
% \end{center}
%    \caption{Histogram of scaled gradients of layer ``conv3'' of ``1-2-6''
%    AlexNet model at epoch 5 and 35. The gradients are scaled to $[-1, 1]$ before
%    being plot.
%    The y-axis is in logarithmic scale.
%    There are 64 possible values since the gradients are 6-bit.}
% \label{fig:gradient_distrib}
% \end{figure}

% %---------------------------------Figure---------------------------------%
% \begin{figure}
% \begin{center}
% \includegraphics[height=0.3\textheight,
% width=0.6\textwidth]{weights}
% \end{center}
%    \caption{Histogram of weights of layer ``conv3'' of ``1-2-6'' AlexNet
%    model at epoch 5, 15 and 35. There are two possible values at a specific
%    epoch since the weights are scaled 1-bit.}
% \label{fig:weight_distrib}
% \end{figure}
% 
% %---------------------------------Figure---------------------------------%
% \begin{figure}
% \begin{center}
% \includegraphics[height=0.3\textheight,
% width=0.6\textwidth]{activations}
% \end{center}
%    \caption{Histogram of activation of layer ``conv3'' of ``1-2-6'' AlexNet
%    model at epoch 5, 15 and 35. There are four possible values at a specific
%    epoch since the activations are 2-bit.}
% \label{fig:activation_distrib}
% \end{figure}

\subsection{Making First and Last Layer low bitwidth}
\label{subsec:first-and-last-layer}
To answer the question whether the first and the last layer need to be treated
specially when quantizing to low bitwidth, we use the same models A, B, C from
Table~\ref{tab:svhn_wag} to find out if it is cost-effective to quantize the
first and last layer to low bitwidth, and collect the results in
Table~\ref{tab:first_and_last}.

It can be seen that quantizing first and the last layer indeed leads to
significant accuracy degradation, and models with less number of channels suffer
more. The degradation % is significant, % on AlexNet,
 to some extent justifies the
practices of BNN and XNOR-net of not quantizing these two layers.

\begin{table}[!ht] \centering \small
\caption{Control experiments for
investigation on theh degredation cost by quantizing the first convolutional layer and the last
FC layer to low bitwidth.
The row with ``(1, 2, 4)'' stands for the baseline case of
$(W,\,A,\,G)=(1,\,2,\,4)$  and not
quantizing the first and last layers.
``+ first'' means additionally quantizing the weights and gradients of the first
convolutional layer (outputs of the first layer are already
quantized in the base ``(1,2,4)'' scheme).
``+ last'' means quantizing the inputs, weights and gradients of the last FC
layer. Note that outputs of the last layer do not need quantization.
%converted to low bitwidth numbers.
}
\begin{center}
\begin{tabular}{
 p{0.25\linewidth} p{0.09\linewidth}
p{0.09\linewidth} p{0.09\linewidth} p{0.09\linewidth}}
\toprule  Scheme & Model A Accuracy & Model B
Accuracy & Model C Accuracy\\% & AlexNet\\
    \midrule (1, 2, 4) & 0.975 & 0.969 & 0.939\\% & 0.471 \\
    \midrule (1, 2, 4) + first & 0.972 & 0.963 & 0.932\\% & 0.425 \\
    \midrule (1, 2, 4) + last & 0.973 & 0.969 & 0.927\\% & 0.403 \\
    \midrule (1, 2, 4) + first + last & 0.971 & 0.961 & 0.928\\% & - \\
\hline
\end{tabular}
\end{center} \label{tab:first_and_last}
\end{table}

\section{Discussion and Related Work}
By binarizing weights and activations, binarized neural networks like BNN and
XNOR-Net have enabled acceleration of the forward pass of neural network with
bit convolution kernel.
However, the backward pass of binarized networks still requires convolutions
between floating-point gradients and weights, which could not efficiently
exploit bit convolution kernel as gradients are in general not low bitwidth
numbers.
%and sensitive to errors.

\citep{lin2015neural} makes a step further towards low bitwidth gradients by
converting some multiplications to bit-shift. However, the number of additions
between high bitwidth numbers remains at the same order of magnitude as before,
leading to reduced overall speedup.

There is also another series of work \citep{seide20141} that quantizes gradients
before communication in distributed computation settings. However, the work is
more concerned with decreasing the amount of communication traffic, and does
not deal with the bitwidth of gradients used in back-propagation. In particular,
they use full precision gradients during the backward pass, and quantize the
gradients only before sending them to other computation nodes. In contrast, we
quantize gradients each time before they reach the selected convolution layers
during the backward pass.

To the best of our knowledge, our work is the first to reduce the
bitwidth of gradient to 6-bit and lower, while still achieving comparable
prediction accuracy without altering other aspects of neural network model, such as
increasing the number of channels, for models as large as AlexNet on ImageNet
dataset.

\section{Conclusion and Future Work}
We have introduced DoReFa-Net, a method to train a convolutional neural network
that has low bitwidth weights and activations using low bitwidth parameter gradients.
We find that weights and activations can be
deterministically quantized while gradients need to be stochastically quantized.

As most
convolutions during forward/backward passes are now taking low bitwidth weights
and activations/gradients respectively, DoReFa-Net can use the bit convolution
kernels to accelerate both training and inference process. Our experiments on SVHN and ImageNet
datasets demonstrate that DoReFa-Net can achieve comparable
prediction accuracy as their 32-bit counterparts. For example, a DoReFa-Net derived
from AlexNet that has 1-bit weights, 2-bit activations, can be trained from
scratch using 6-bit gradients to get 46.1\% top-1 accuracy on ImageNet
validation set.

As future work, it would be interesting to investigate using FPGA to train
DoReFa-Net, as the $O(B^2)$ resource requirement of computation units for
$B$-bit arithmetic on FPGA strongly favors low bitwidth convolutions.

%\bibliographystyle{spmpsci}
%\bibliography{thesis}
%%\bibliographystyle{plain}
%%\input{goo.bbl}

%\end{document}

%\maketitle

%\begin{abstract}
%The abstract paragraph should be indented 1/2~inch (3~picas) on both left and
%right-hand margins. Use 10~point type, with a vertical spacing of 11~points.
%The word \textsc{Abstract} must be centered, in small caps, and in point size 12. Two
%line spaces precede the abstract. The abstract must be limited to one
%paragraph.
%\end{abstract}
%\input{document}

\bibliography{thesis}

\begin{thebibliography}{28}
\providecommand{\natexlab}[1]{#1}
\providecommand{\url}[1]{\texttt{#1}}
\expandafter\ifx\csname urlstyle\endcsname\relax
  \providecommand{\doi}[1]{doi: #1}\else
  \providecommand{\doi}{doi: \begingroup \urlstyle{rm}\Url}\fi

\bibitem[Abadi et~al.()Abadi, Agarwal, Barham, Brevdo, Chen, Citro, Corrado,
  Davis, Dean, Devin, et~al.]{abaditensorflow}
Abadi, Mart{\i}n, Agarwal, Ashish, Barham, Paul, Brevdo, Eugene, Chen, Zhifeng,
  Citro, Craig, Corrado, Greg~S, Davis, Andy, Dean, Jeffrey, Devin, Matthieu,
  et~al.
\newblock Tensorflow: Large-scale machine learning on heterogeneous systems,
  2015.
\newblock \emph{Software available from tensorflow. org}.

\bibitem[Bahdanau et~al.(2014)Bahdanau, Cho, and Bengio]{bahdanau2014neural}
Bahdanau, Dzmitry, Cho, Kyunghyun, and Bengio, Yoshua.
\newblock Neural machine translation by jointly learning to align and
  translate.
\newblock \emph{arXiv preprint arXiv:1409.0473}, 2014.

\bibitem[Bengio et~al.(2013)Bengio, L{\'e}onard, and
  Courville]{bengio2013estimating}
Bengio, Yoshua, L{\'e}onard, Nicholas, and Courville, Aaron.
\newblock Estimating or propagating gradients through stochastic neurons for
  conditional computation.
\newblock \emph{arXiv preprint arXiv:1308.3432}, 2013.

\bibitem[Chen et~al.(2014{\natexlab{a}})Chen, Du, Sun, Wang, Wu, Chen, and
  Temam]{chen2014diannao}
Chen, Tianshi, Du, Zidong, Sun, Ninghui, Wang, Jia, Wu, Chengyong, Chen, Yunji,
  and Temam, Olivier.
\newblock Diannao: A small-footprint high-throughput accelerator for ubiquitous
  machine-learning.
\newblock In \emph{ACM Sigplan Notices}, volume~49, pp.\  269--284. ACM,
  2014{\natexlab{a}}.

\bibitem[Chen et~al.(2014{\natexlab{b}})Chen, Luo, Liu, Zhang, He, Wang, Li,
  Chen, Xu, Sun, et~al.]{chen2014dadiannao}
Chen, Yunji, Luo, Tao, Liu, Shaoli, Zhang, Shijin, He, Liqiang, Wang, Jia, Li,
  Ling, Chen, Tianshi, Xu, Zhiwei, Sun, Ninghui, et~al.
\newblock Dadiannao: A machine-learning supercomputer.
\newblock In \emph{Proceedings of the 47th Annual IEEE/ACM International
  Symposium on Microarchitecture}, pp.\  609--622. IEEE Computer Society,
  2014{\natexlab{b}}.

\bibitem[Courbariaux \& Bengio(2016)Courbariaux and
  Bengio]{courbariaux2016binarynet}
Courbariaux, Matthieu and Bengio, Yoshua.
\newblock Binarynet: Training deep neural networks with weights and activations
  constrained to+ 1 or-1.
\newblock \emph{arXiv preprint arXiv:1602.02830}, 2016.

\bibitem[Courbariaux et~al.(2014)Courbariaux, Bengio, and
  David]{courbariaux2014training}
Courbariaux, Matthieu, Bengio, Yoshua, and David, Jean-Pierre.
\newblock Training deep neural networks with low precision multiplications.
\newblock \emph{arXiv preprint arXiv:1412.7024}, 2014.

\bibitem[Deng et~al.(2009)Deng, Dong, Socher, Li, Li, and
  Fei-Fei]{deng2009imagenet}
Deng, Jia, Dong, Wei, Socher, Richard, Li, Li-Jia, Li, Kai, and Fei-Fei, Li.
\newblock Imagenet: A large-scale hierarchical image database.
\newblock In \emph{Computer Vision and Pattern Recognition, 2009. CVPR 2009.
  IEEE Conference on}, pp.\  248--255. IEEE, 2009.

\bibitem[Farabet et~al.(2011)Farabet, LeCun, Kavukcuoglu, Culurciello, Martini,
  Akselrod, and Talay]{farabet2011large}
Farabet, Cl{\'e}ment, LeCun, Yann, Kavukcuoglu, Koray, Culurciello, Eugenio,
  Martini, Berin, Akselrod, Polina, and Talay, Selcuk.
\newblock Large-scale fpga-based convolutional networks.
\newblock \emph{Scaling up Machine Learning: Parallel and Distributed
  Approaches}, pp.\  399--419, 2011.

\bibitem[Gong et~al.(2014)Gong, Liu, Yang, and Bourdev]{gong2014compressing}
Gong, Yunchao, Liu, Liu, Yang, Ming, and Bourdev, Lubomir.
\newblock Compressing deep convolutional networks using vector quantization.
\newblock \emph{arXiv preprint arXiv:1412.6115}, 2014.

\bibitem[Gupta et~al.(2015)Gupta, Agrawal, Gopalakrishnan, and
  Narayanan]{gupta2015deep}
Gupta, Suyog, Agrawal, Ankur, Gopalakrishnan, Kailash, and Narayanan, Pritish.
\newblock Deep learning with limited numerical precision.
\newblock \emph{arXiv preprint arXiv:1502.02551}, 2015.

\bibitem[Han et~al.(2015{\natexlab{a}})Han, Mao, and Dally]{han2015deep}
Han, Song, Mao, Huizi, and Dally, William~J.
\newblock Deep compression: Compressing deep neural networks with pruning,
  trained quantization and huffman coding.
\newblock \emph{arXiv preprint arXiv:1510.00149}, 2015{\natexlab{a}}.

\bibitem[Han et~al.(2015{\natexlab{b}})Han, Pool, Tran, and
  Dally]{han2015learning}
Han, Song, Pool, Jeff, Tran, John, and Dally, William.
\newblock Learning both weights and connections for efficient neural network.
\newblock In \emph{Advances in Neural Information Processing Systems}, pp.\
  1135--1143, 2015{\natexlab{b}}.

\bibitem[Hinton et~al.(2012{\natexlab{a}})Hinton, Deng, Yu, Dahl, Mohamed,
  Jaitly, Senior, Vanhoucke, Nguyen, Sainath, et~al.]{hinton2012deep}
Hinton, Geoffrey, Deng, Li, Yu, Dong, Dahl, George~E, Mohamed, Abdel-rahman,
  Jaitly, Navdeep, Senior, Andrew, Vanhoucke, Vincent, Nguyen, Patrick,
  Sainath, Tara~N, et~al.
\newblock Deep neural networks for acoustic modeling in speech recognition: The
  shared views of four research groups.
\newblock \emph{Signal Processing Magazine, IEEE}, 29\penalty0 (6):\penalty0
  82--97, 2012{\natexlab{a}}.

\bibitem[Hinton et~al.(2012{\natexlab{b}})Hinton, Srivastava, and
  Swersky]{hinton2012neural}
Hinton, Geoffrey, Srivastava, Nitsh, and Swersky, Kevin.
\newblock Neural networks for machine learning.
\newblock \emph{Coursera, video lectures}, 264, 2012{\natexlab{b}}.

\bibitem[Ioffe \& Szegedy(2015)Ioffe and Szegedy]{ioffe2015batch}
Ioffe, Sergey and Szegedy, Christian.
\newblock Batch normalization: Accelerating deep network training by reducing
  internal covariate shift.
\newblock \emph{arXiv preprint arXiv:1502.03167}, 2015.

\bibitem[Kim \& Smaragdis(2016)Kim and Smaragdis]{kim2016bitwise}
Kim, Minje and Smaragdis, Paris.
\newblock Bitwise neural networks.
\newblock \emph{arXiv preprint arXiv:1601.06071}, 2016.

\bibitem[Kingma \& Ba(2014)Kingma and Ba]{kingma2014adam}
Kingma, Diederik and Ba, Jimmy.
\newblock Adam: A method for stochastic optimization.
\newblock \emph{arXiv preprint arXiv:1412.6980}, 2014.

\bibitem[Krizhevsky et~al.(2012)Krizhevsky, Sutskever, and
  Hinton]{krizhevsky2012imagenet}
Krizhevsky, Alex, Sutskever, Ilya, and Hinton, Geoffrey~E.
\newblock Imagenet classification with deep convolutional neural networks.
\newblock In \emph{Advances in neural information processing systems}, pp.\
  1097--1105, 2012.

\bibitem[Li \& Liu(2016)Li and Liu]{li2016ternary}
Li, Fengfu and Liu, Bin.
\newblock Ternary weight networks.
\newblock \emph{arXiv preprint arXiv:1605.04711}, 2016.

\bibitem[Lin et~al.(2015)Lin, Courbariaux, Memisevic, and
  Bengio]{lin2015neural}
Lin, Zhouhan, Courbariaux, Matthieu, Memisevic, Roland, and Bengio, Yoshua.
\newblock Neural networks with few multiplications.
\newblock \emph{arXiv preprint arXiv:1510.03009}, 2015.

\bibitem[Merolla et~al.(2016)Merolla, Appuswamy, Arthur, Esser, and
  Modha]{merolla2016deep}
Merolla, Paul, Appuswamy, Rathinakumar, Arthur, John, Esser, Steve~K, and
  Modha, Dharmendra.
\newblock Deep neural networks are robust to weight binarization and other
  non-linear distortions.
\newblock \emph{arXiv preprint arXiv:1606.01981}, 2016.

\bibitem[Netzer et~al.(2011)Netzer, Wang, Coates, Bissacco, Wu, and
  Ng]{netzer2011reading}
Netzer, Yuval, Wang, Tao, Coates, Adam, Bissacco, Alessandro, Wu, Bo, and Ng,
  Andrew~Y.
\newblock Reading digits in natural images with unsupervised feature learning.
\newblock In \emph{NIPS workshop on deep learning and unsupervised feature
  learning}, volume 2011, pp.\ ~5. Granada, Spain, 2011.

\bibitem[Pham et~al.(2012)Pham, Jelaca, Farabet, Martini, LeCun, and
  Culurciello]{pham2012neuflow}
Pham, Phi-Hung, Jelaca, Darko, Farabet, Clement, Martini, Berin, LeCun, Yann,
  and Culurciello, Eugenio.
\newblock Neuflow: Dataflow vision processing system-on-a-chip.
\newblock In \emph{Circuits and Systems (MWSCAS), 2012 IEEE 55th International
  Midwest Symposium on}, pp.\  1044--1047. IEEE, 2012.

\bibitem[Rastegari et~al.(2016)Rastegari, Ordonez, Redmon, and
  Farhadi]{rastegari2016xnor}
Rastegari, Mohammad, Ordonez, Vicente, Redmon, Joseph, and Farhadi, Ali.
\newblock Xnor-net: Imagenet classification using binary convolutional neural
  networks.
\newblock \emph{arXiv preprint arXiv:1603.05279}, 2016.

\bibitem[Seide et~al.(2014)Seide, Fu, Droppo, Li, and Yu]{seide20141}
Seide, Frank, Fu, Hao, Droppo, Jasha, Li, Gang, and Yu, Dong.
\newblock 1-bit stochastic gradient descent and its application to
  data-parallel distributed training of speech dnns.
\newblock In \emph{INTERSPEECH}, pp.\  1058--1062, 2014.

\bibitem[Vanhoucke et~al.(2011)Vanhoucke, Senior, and
  Mao]{vanhoucke2011improving}
Vanhoucke, Vincent, Senior, Andrew, and Mao, Mark~Z.
\newblock Improving the speed of neural networks on cpus.
\newblock In \emph{Proc. Deep Learning and Unsupervised Feature Learning NIPS
  Workshop}, volume~1, 2011.

\bibitem[Wu et~al.(2015)Wu, Leng, Wang, Hu, and Cheng]{wu2015quantized}
Wu, Jiaxiang, Leng, Cong, Wang, Yuhang, Hu, Qinghao, and Cheng, Jian.
\newblock Quantized convolutional neural networks for mobile devices.
\newblock \emph{arXiv preprint arXiv:1512.06473}, 2015.

\end{thebibliography}
\bibliographystyle{iclr2016_conference}

\end{document}